\documentclass[conference,a4paper]{IEEEtran}
\IEEEoverridecommandlockouts

\usepackage[hidelinks]{hyperref}
\usepackage[cmex10]{amsmath}%
\usepackage{amssymb,amsfonts}%
\usepackage{dblfloatfix}%

\usepackage[ruled,vlined]{algorithm2e}
\usepackage{graphicx}
\graphicspath{{Figures/PDF/}{Figures/PNG/}}

\usepackage[numbers,compress]{natbib}
\usepackage{texnames}
\usepackage{bm,bbm}
\usepackage{orcidlink}
\usepackage{subcaption}
\usepackage{xcolor}
\usepackage{array}
\usepackage{multirow}
\usepackage{balance}
\newcolumntype{C}[1]{>{\centering\let\newline\\\arraybackslash\hspace{0pt}}m{#1}}
\usepackage{float}
\usepackage{hyperref}
\usepackage{tabularx}
\usepackage{placeins}

\begin{document}

\title{\fontsize{14}{16}\selectfont DUAL-DOMAIN MASKED IMAGE MODELING: A SELF-SUPERVISED PRETRAINING STRATEGY USING SPATIAL AND FREQUENCY DOMAIN MASKING FOR HYPERSPECTRAL DATA
}

\author{
    \IEEEauthorblockN{
        Shaheer Mohamed$^{1,2}$, 
        Tharindu Fernando$^1$, 
        Sridha Sridharan$^1$, 
        Peyman Moghadam$^{2,1}$, 
        Clinton Fookes$^1$
    }
    
    \vspace{0.5em} %

    \IEEEauthorblockA{
        $^1$Signal Processing, Artificial Intelligence and Vision Technologies, Queensland University of Technology, Brisbane, Australia \\
        $^2$CSIRO Robotics, Data61, CSIRO, Brisbane, QLD, Australia
    }
}

\maketitle
\begin{abstract}
Hyperspectral images (HSIs) capture rich spectral signatures that reveal vital material properties, offering broad applicability across various domains. However, the scarcity of labeled HSI data limits the full potential of deep learning, especially for transformer-based architectures that require large-scale training. To address this constraint, we propose Spatial-Frequency Masked Image Modeling (SFMIM), a self-supervised pretraining strategy for hyperspectral data that utilizes the large portion of unlabeled data. Our method introduces a novel dual-domain masking mechanism that operates in both spatial and frequency domains. The input HSI cube is initially divided into non-overlapping patches along the spatial dimension, with each patch comprising the entire spectrum of its corresponding spatial location. In spatial masking, we randomly mask selected patches and train the model to reconstruct the masked inputs using the visible patches. Concurrently, in frequency masking, we remove portions of the frequency components of the input spectra and predict the missing frequencies. By learning to reconstruct these masked components, the transformer-based encoder captures higher-order spectral–spatial correlations. We evaluate our approach on three publicly available HSI classification benchmarks and demonstrate that it achieves state-of-the-art performance. Notably, our model shows rapid convergence during fine-tuning, highlighting the efficiency of our pretraining strategy.
\end{abstract}
\vspace{2.5mm}

\section{Introduction}
Hyperspectral images (HSIs) capture detailed spectral information across a broad range of the electromagnetic spectrum. These rich spectral signatures encode crucial material properties, enabling their use in numerous applications, such as precision agriculture \cite{moghadam2017plant}, environmental monitoring \cite{Spatioformer}, medical diagnostics \cite{medical}, food quality analysis \cite{food-quality}, and defence or security tasks \cite{military}, where perception beyond the visible spectrum is essential. Crucially, hyperspectral sensors collect both spatial and spectral information, and these two facets complement each other to comprehensively represent a scene.

Early works in hyperspectral image analysis predominantly employed classical machine learning techniques~\cite{SVM, knn-song2016hyperspectral}, which were later improved by convolutional neural networks (CNNs)~\cite{1D-CNN, 2D-CNN, miniGCN}.
While CNN-based backbones remain the foundation for most prior approaches, they exhibit limitations in modeling long-range dependencies \cite{vit_original, hong2021spectralformer}. %
Recently, transformer architectures have emerged as the state-of-the-art in computer vision and natural language processing, owing to their ability to effectively capture such dependencies. This paradigm shift has also begun to influence the hyperspectral imaging domain \cite{Spatioformer, mohamed2023factoformer}.

Despite these developments, limited labeled hyperspectral datasets continue to hinder the full exploitation of transformer models. In many public datasets, only a small portion of the available data is labeled, due to the expense and complexity of manual labeling. Self-supervised learning addresses this issue by learning representations directly from unlabeled data through various pretext tasks \cite{SimMIM, MAE}. A handful of studies have applied this idea to hyperspectral imagery, MAEST \cite{MAEST} employs masked image modeling along the spectral dimension, while FactoFormer \cite{mohamed2023factoformer} introduces a factorized transformer that separately learns spatial and spectral features using tailored masking schemes to pre-train both branches. However, these approaches does not explore the incorporation of frequency-domain information during pre-training.

To overcome these limitations, we propose Spatial-Frequency Masked Image Modeling (SFMIM), a dual-domain masking strategy for hyperspectral pretraining. Specifically, we recognize that hyperspectral images are composed of two inherently coupled dimensions: the spatial domain, across 2D image coordinates, and the spectral domain, across wavelength or frequency bands. Our framework applies spatial and frequency domain masking to fully exploit these joint spatial–spectral correlations: 1) Spatial Masking: we randomly mask entire patches in the spatial domain, replacing them with trainable mask tokens. This forces the network to infer missing spatial information from the remaining unmasked patches, thereby learning robust local and global spatial relationships across different spectral channels. 2) Frequency Domain Masking: in parallel, we corrupt the spectral dimension by applying a Fourier transform and selectively filtering out (masking) certain frequency components. By forcing the network to predict the missing frequency components, we enable it to learn salient spectral features. 

This dual-domain training scheme enables the model to learn highly discriminative features without relying on large labeled datasets. In practice, we leverage abundant unlabeled hyperspectral data for pretraining, then fine-tune on limited labeled subsets, significantly reducing data annotation requirements. Empirical results on multiple publicly available datasets confirm that our approach not only achieves state-of-the-art performance but also converges rapidly during fine-tuning, highlighting the effectiveness of representation learning during pretraining.

The key contributions of the paper are summarized as follows:
\begin{itemize}
        \item We propose a novel self-supervised pretraining strategy for hyperspectral transformers, enabling the utilization of a large proportion of unlabeled data and unlocking the full potential of transformer networks.
        \item We introduce a novel dual-domain masking strategy, SFMIM, where the input is masked in both spatial and frequency domains. This masked image modeling-based pretraining allows the model to learn robust spatial-spectral representations inherent to hyperspectral data.
        \item We evaluate the proposed method using three publicly available benchmarks for HSI classification. Our method achieves state-of-the-art performance, outperforming both supervised and self-supervised transformer-based methods, while demonstrating rapid convergence and reduced computational complexity.    
\end{itemize}

\section{Related Works}
With the success of transformers and masked image modeling in computer vision domain \cite{SimMIM,MAE,xie2023masked, haghighat2024pre}, their application to hyperspectral imaging has grown, surpassing traditional CNN-based models~\cite{hong2021spectralformer, mohamed2023factoformer, MAEST, scheibenreif2023masked, shaheer2024tracking}. In the hyperspectral domain, some works have explored hybrid models, which combine transformers and CNNs, while more recent works focus on designing different versions of transformer networks specifically for hyperspectral data. SpectralFormer \cite{hong2021spectralformer} introduced a fully transformer-based network for HSI classification, incorporating an additional skip connection. However, since this approach was entirely supervised and hyperspectral data are limited, the full potential of transformer-based networks was not fully realized. 

Subsequently, with masked image modeling emerging as a successful self-supervised pretraining method in computer vision, MAEST \cite{MAEST} adopted masked image pretraining for hyperspectral data. MAEST proposed a masked autoencoder architecture inspired by MAE \cite{MAE}, where the input is patched only along the spectral dimension, and binary masking is applied  along the spectral dimension where spatial correlations were not fully realized. MaskedSST \cite{scheibenreif2023masked} proposed masked modeling by patching the input in both spatial and spectral dimensions and applying random masking. Additionally, they introduced spatial-spectral factorized attention to process each dimension independently. FactoFormer \cite{mohamed2023factoformer} introduced a factorized transformer architecture that processes the spatial and spectral dimensions separately using two distinct transformers, which are fused at the final stage. Additionally, they pre-train the spatial and spectral transformers independently using masked image modeling. However, employing separate transformers for spatial and spectral processing increases computational complexity. Moreover, the independent pretraining of spatial and spectral dimensions limits the model’s ability to capture higher-order spatial-spectral correlations.

We propose SFMIM, a dual-domain masking strategy that masks the input in both spatial and frequency domains to capture spatial and spectral features simultaneously. Compared to MAEST \cite{MAEST}, our approach incorporates masking in both spatial and spectral domains, and we emphasize that our method uses frequency-based masking for spectra instead of random binary masking. Additionally, compared to FactoFormer \cite{mohamed2023factoformer}, our method is more computationally efficient, as it uses a single transformer to process both spatial and spectral features. Furthermore, our pretraining effectively captures spatial-spectral correlations simultaneously, whereas FactoFormer requires separate pretraining for the spatial and spectral dimensions.

\section{SFMIM: Dual-Domain Masked Image Modeling}

\begin{figure*}[t!]
    \centering
    \includegraphics[scale=0.85]{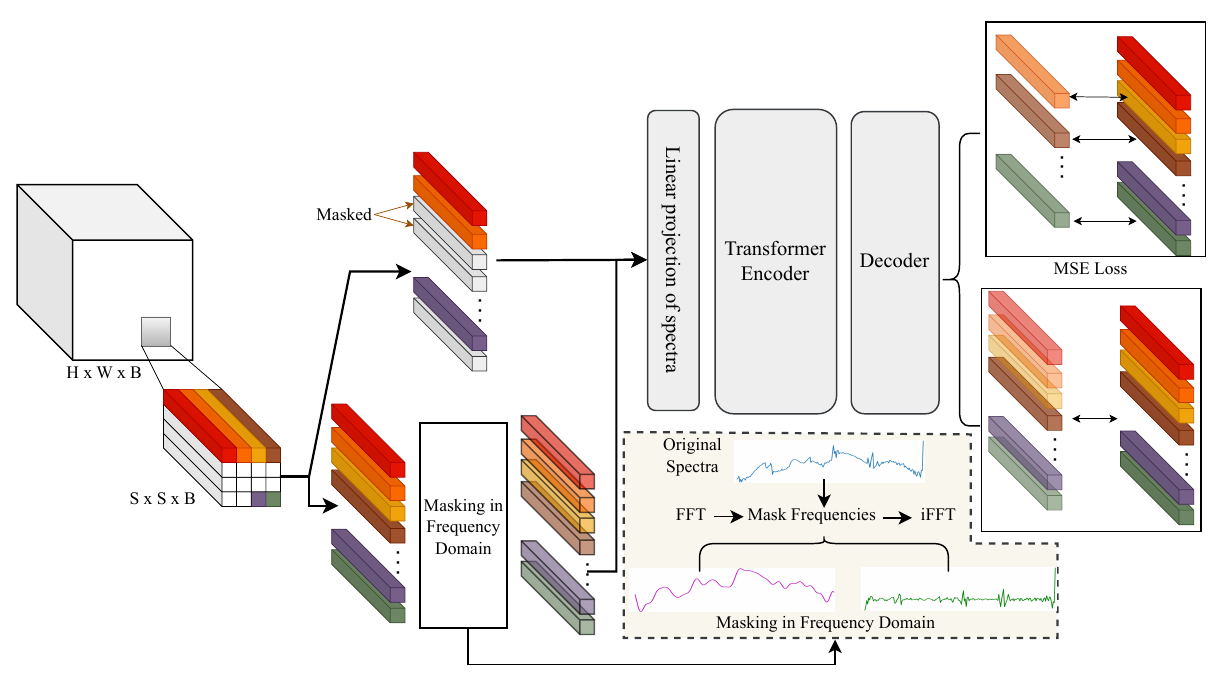}
    \vspace{-2mm}
    \caption{An overview of the proposed SFMIM network architecture. The input hyperspectral cube is split into non-overlapping patches along the spatial dimension, masked in both spatial and frequency domains, and processed by a transformer encoder. A lightweight decoder reconstructs the masked input with MSE used as the reconstruction loss.}
    \label{fig:overview}
    \vspace{-5mm}
\end{figure*}

In this section, we present our proposed framework, which leverages a transformer-based backbone and employs masked image modeling-based pretraining tailored for hyperspectral data. By applying masking in both spatial and frequency domains, our approach efficiently learns the spatial-spectral features of the hyperspectral data. The overall network architecture is illustrated in Fig.\ref{fig:overview}.

\subsection{Transformer-Based Encoder}
For each hyperspectral cube, we select an extended neighborhood of size $S \times S$. Let the input be $X \in \mathbb{R}^{B \times S \times S}$, where $B$ denotes the number of spectral bands, and $(S, S)$ represents the spatial dimensions. We divide $X$ into $N = S^2$ non-overlapping patches spatially, where each patch is defined as $y_i \in \mathbb{R}^{B}$, capturing the complete spectral vector for each individual spatial location. The resulting sequence of patches is then fed into the transformer network.

We adhere to the standard Vision Transformer architecture, utilizing multi-head self-attention as described in~\cite{vit_original}. Each spectral vector $y_i \in \mathbb{R}^{B}$ is projected into an embedding space of dimension $d$ via a linear layer $\mathbf{E} \in \mathbb{R}^{d \times B}$. Additionally, a positional embedding vector $\mathbf{e}_i \in \mathbb{R}^d$ is added to incorporate positional information. This process is formalized in Eq.~\ref{eq:sproj}:

\begin{equation}
    z_i = \mathbf{E} \,(y_i) + \mathbf{e}_i, \quad i=1,\dots,N.
    \label{eq:sproj}
\end{equation}

We prepend a learnable classification token $z_{cls} \in \mathbb{R}^d$ to the sequence $\{z_1, z_2, \dots, z_N\}$. Hence, the final input to the transformer based encoder is,  
\begin{equation}
    Z_{enc} = [z_{cls}, z_1, z_2, \dots, z_N] \in \mathbb{R}^{(N+1)\times d}
    \label{eq:seq}
\end{equation}
where $z_{cls}$ in Eq. \ref{eq:seq} serves as a global representation of the entire patch sequence.

\noindent \textbf{Spatial Masking:} We randomly apply a binary mask $M \in \{0,1\}^{N}$ to each spatial patch, indicating which spectral vectors are removed and replaced with learnable masked tokens. Here, each patch is of spatial size $1 \times 1 \times B$, meaning that each flattened patch captures the full spectrum of a single pixel. The model processes only a small subset of visible tokens and reconstructs the masked spectra using a lightweight linear decoder. The model is trained using a Mean Squared Error (MSE) loss. By inferring the missing spectra from neighboring unmasked tokens, the model leverages local spatial-spectral context, effectively capturing subtle variations across adjacent pixels.

\vspace{2.5mm}

\noindent \textbf{Frequency Masking:} For each spatial location $(h,w)$, we treat the corresponding spectral vector $x_{h,w} \in \mathbb{R}^B$ as a 1D signal and compute its discrete Fourier transform $ \tilde{\mathbf{x}}_{h, w}$ using Eq.\ref{eq:fft},

\begin{equation}
    \tilde{\mathbf{x}}_{h, w} = \mathcal{F}(\mathbf{x}_{h, w}) \in \mathbb{C}^{\lceil B/2 \rceil + 1},  
    \label{eq:fft}
\end{equation}
where $\mathcal{F}$ denotes the Fourier transform operator defined in Eq.~\ref{eq:fourier},

\begin{equation}
\mathcal{F}(\mathbf{x_{h,w}})(k) = \sum_{n=0}^{B-1} \mathbf{x}_{h, w}(n) \cdot e^{-i \, 2\pi \frac{k n}{B}}, \quad k = 0, 1, \dots, B-1,
\label{eq:fourier}
\end{equation}
where $ \mathbf{x}_{h, w}(n)$ is the real-valued spectral intensity in the $n^{th}$ band and $\mathcal{F}(\mathbf{x_{h,w}})(k)$ is the complex-valued frequency component at the $k^{th}$ frequency. 

We apply low-pass and high-pass filters to remove specific frequency information from each input cube. Specifically, for each spectral vector $\tilde{\mathbf{x}}_{h,w}$ in the frequency domain, we randomly choose to apply either a low-pass or high-pass filter with a cutoff frequency $\alpha = \gamma \lceil B/2 \rceil$, where $0 < \gamma < 1$. The masked frequency components are defined as:

\begin{equation}
    \tilde{\mathbf{x}}^{\text{mask}}_{h, w}[k] =
\begin{cases} 
0, & \text{if low-pass and } k > \alpha, \\
0, & \text{if high-pass and } k \leq \alpha, \\
\tilde{\mathbf{x}}_{h, w}[k], & \text{otherwise}.
\end{cases}
\label{eq:condition}
\end{equation}

After applying the frequency mask in the frequency domain, we reconstruct the original spectral signal by performing an inverse Fourier transform, as shown in Eq.\ref{eq:inverse_fft},

\begin{equation}
   \mathbf{x}_{h, w}^{\text{mask}} = \mathcal{F}^{-1} \big( \tilde{\mathbf{x}}_{h, w}^{\text{mask}} \big) \in \mathbb{R}^C ,
   \label{eq:inverse_fft}
\end{equation}
where, $\mathcal{F}^{-1}(\mathbf{x_{h,w}})$ defines the inverse Fourier transform that maps the frequency-domain vector back to the spectral domain. 
The frequency-masked input sequence, $X^{\text{mask}} \in \mathbb{R}^{B \times N}$, is fed into the transformer layers after passing through the patch embedding, adding positional embeddings, and appending the classification token, as previously described in Sec. 3.A. We employ the same lightweight linear decoder to reconstruct the original spectral vectors from the frequency-masked inputs and train the model using an MSE loss.

\vspace{-1mm}

\section{Experiments}
\vspace{-1mm}
\begin{table*}[t]
\centering
\caption{Quantitative Performance of Different Classification Methods in Terms of OA, AA, and k on Indian Pines, University of Pavia, and Houston 2013 Datasets. The Best Results are Bold, and the Second Best are Underlined.}
\label{tab:results}
\resizebox{\textwidth}{!}{%
\begin{tabular}{c|c|cccc|ccccc}
\hline 
Dataset & Metric & \multicolumn{4}{c|}{Classic Backbone Networks} & \multicolumn{5}{c}{Transformers} \\
\hline
\multicolumn{2}{c|}{} & \multicolumn{1}{c}{1-D CNN~\cite{1D-CNN}} & \multicolumn{1}{c}{2-D CNN~\cite{2D-CNN}} & \multicolumn{1}{c}{RNN~\cite{RNN}} & \multicolumn{1}{c|}{miniGCN~\cite{miniGCN}} & \multicolumn{1}{c}{ViT~\cite{vit_original}} & \multicolumn{1}{c}{SpectralFormer~\cite{hong2021spectralformer}} & \multicolumn{1}{c}{MAEST~\cite{MAEST}} & \multicolumn{1}{c}{FactoFormer~\cite{mohamed2023factoformer}} & \multicolumn{1}{c}{\textbf{SFMIM} (Ours)} \\
\hline
\hline 
\multirow{3}{*}{Indian Pines} & OA (\%) & 70.43 & 75.89 & 70.66 & 75.11 & 71.86 & 81.76 & 84.15 & \textbf{91.30} & \underline{90.23} \\
 & AA (\%) & 79.6 & 86.64 & 76.37 & 78.03 & 78.97 & 87.81 & 90.97 & \underline{94.30} & \textbf{94.45} \\
 & \textit{k} & 0.6642 & 0.7281 & 0.6673 & 0.7164 & 0.6804 & 0.7919 & 0.8200 & \textbf{0.9006} & \underline{0.8883} \\
\hline
\multirow{3}{*}{University of Pavia} & OA (\%) & 75.5 & 86.05 & 77.13 & 79.79 & 76.99 & 91.07 & 91.06 & \textbf{95.19} & \underline{94.57} \\
 & AA (\%) & 86.26 & 88.99 & 84.29 & 85.07 & 80.22 & 90.2 & 90.00 & \textbf{93.64} & \underline{92.72} \\
 & \textit{k} & 0.6948 & 0.8187 & 0.7101 & 0.7367 & 0.7010 & 0.8805 & 0.8794 & \textbf{0.9349} & \underline{0.9263} \\
\hline
\multirow{3}{*}{Houston 2013} & OA (\%) & 80.04 & 83.72 & 83.23 & 81.71 & 80.41 & 88.01 & 88.55 & \underline{89.13} & \textbf{91.15} \\
 & AA (\%) & 82.74 & 84.35 & 85.04 & 83.09 & 82.50 & 88.91 & 88.89 & \underline{90.12} & \textbf{92.43} \\
 & \textit{k} & 0.7835 & 0.8231 & 0.8183 & 0.8018 & 0.7876 & 0.8699 & 0.8757 & \underline{0.8820} & \textbf{0.9040} \\
\hline 
\end{tabular}%
}
\vspace{-3mm}
\end{table*}
In this section, we present our experimental setup, the evaluated datasets, and implementation details. We also provide results comparing our method with other transformer-based networks and classical backbones. Additionally, we present a comprehensive analysis of different masking strategies used for hyperspectral transformer pretraining.

 \subsection{Experimental Setting}

We conducted experiments on three widely recognized hyperspectral datasets: Indian Pines, University of Pavia, and Houston 2013, utilizing the same training and testing splits as specified in \cite{mohamed2023factoformer}. For performance evaluation, we employed three standard metrics: Overall Accuracy (OA), Average Accuracy (AA), and the Kappa Coefficient, adhering to established evaluation protocols.

The transformer backbone in our architecture consists of five layers, each with four attention heads and an embedding size of 64, consistent with the configurations used in \cite{hong2021spectralformer, mohamed2023factoformer}. We utilized the unlabeled portions of each dataset to pretrain the network, setting the pretraining for 200 epochs. We set the spatial masking ratio to 0.7 and the frequency cut-off $\gamma$ to 0.3, based on empirical testing. We used the pretrained encoder weights and performed end-to-end finetuning on each dataset using the limited labeled samples for the classification task. The fine-tuning models converged in 35 epochs for Indian Pines, 25 epochs for the University of Pavia, and 20 epochs for the Houston 2013 dataset.

\vspace{-2mm}

\subsection{Results and Discussion}
We compare our method with state-of-the-art transformer networks and classical backbone networks. The overall results are shown in Table \ref{tab:results}. In general, transformer based networks perform better compared to classical CNN-based networks. Compared to the fully supervised transformer approach SpectralFormer \cite{hong2021spectralformer}, our method achieved 8.47\%, 3.5\% and 3.14\% higher accuracy on Indian Pines, University of Pavia and Houston 2013 datasets, respectively. This shows the effectiveness of using self-supervised pretraining with transformer-based networks. Also, our proposed method outperforms MAEST \cite{MAEST}, by 6.08\% on Indian Pines, 3.51\% on University of Pavia, and 2.6\% on Houston 2013 datasets respectively. MAEST \cite{MAEST} is a fully transformer network that utilizes masked image modelling base pretraining. This improvement shows the effectiveness of our pretraining approach.

We also compare our model with FactoFormer \cite{mohamed2023factoformer}, a state-of-the-art self-supervised transformer-based approach. Our model outperforms FactoFormer by 2.02\% on the Houston 2013 dataset and achieves comparable performance on the other two datasets, with a margin of approximately 1\%. Importantly, our model not only demonstrates higher accuracy but also greater computational efficiency, especially when compared with FactoFormer. While FactoFormer employs two separate transformers to process spatial and spectral information independently, our approach utilizes a single transformer. FactoFormer requires 80 epochs to converge on both the Indian Pines and University of Pavia datasets and 40 epochs on the Houston 2013 dataset. In contrast, our model converges in only 35, 25, and 20 epochs on the Indian Pines, University of Pavia, and Houston 2013 datasets, respectively. This reduction in convergence time underscores the computational efficiency and highlights the effectiveness of the proposed pertaining strategy. Furthermore, FactoFormer requires separate pre-training for the spatial and spectral transformers. In contrast, our method integrates spatial and frequency domain masking into a single pipeline, enabling the simultaneous learning of inherent spatial and spectral features in hyperspectral data within a single training run. This unified training process not only simplifies the training process but also enhances the model's ability to capture complex spatial-spectral features.

\vspace{-2mm}
\subsection{Effect of Dual-Domain Masking }

Furthermore, to highlight the impact of dual-domain masking, we analyze how different masking strategies perform compared to our method. We compare the models by applying spatial and frequency masking individually. Additionally, we also compare spectral masking as implemented in MAEST \cite{MAEST} and the combined spatial and spectral masking approach in FactoFormer \cite{mohamed2023factoformer}. We evaluate the performance on Houston 2013 dataset, a comparatively large and complex. The results are shown in Table \ref{tab:masking_methods}. Our proposed dual-domain masking outperforms all other methods with significant margins, and converges in only 25 epochs. This shows that the combination of spatial and frequency domain masking learns robust representation of hyperspectral data.

\begin{table}[t]
\centering
\caption{Analysis on Different Masking Strategies for Hyperspectral Transformer Pretraining.}
\label{tab:masking_methods}
\resizebox{\linewidth}{!}{%
\begin{tabular}{c|c|c|c|c|c}
\hline 
Metric & \begin{tabular}[c]{@{}c@{}}Spatial \\ Masking\end{tabular} & \begin{tabular}[c]{@{}c@{}} Spectral  \\Masking \end{tabular} & \begin{tabular}[c]{@{}c@{}}Frequency\\ Masking\\ \end{tabular} & \begin{tabular}[c]{@{}c@{}} Spatial \& Spectral\\ Masking\\ \end{tabular} & \begin{tabular}[c]{@{}c@{}} Dual-Domain\\ Masking\\ \end{tabular} \\
\hline
OA (\%) & 87.30 & 88.55 & 85.03 & 89.13 & \textbf{91.15} \\
AA (\%) & 89.48 & 88.89 & 87.70 & 90.12 & \textbf{92.43} \\
\textit{k} & 0.8623 & 0.8757 & 0.8376 & 0.8820 & \textbf{0.904} \\
\hline 
\end{tabular}%
}
\vspace{-5mm}
\end{table}

\vspace{-1.5mm}

\section{Conclusion}
In this work, we propose SFMIM, a dual-domain masking strategy for pretraining transformer-based models on hyperspectral data. Our approach involves masking the input in both spatial and frequency domains and training the model to reconstruct the original input. Compared to other masked modeling approaches, our method achieves significant performance improvements with rapid convergence during fine-tuning, demonstrating the efficiency of our pretraining strategy. Moreover, employing spatial masking to learn spatial features and frequency masking to extract spectral features within a unified pipeline enables the network to learn robust spatial-spectral representations from hyperspectral data.

\balance{}
\small
\bibliographystyle{IEEEtranN}
\bibliography{references}

\end{document}